\documentclass[letterpaper, 10 pt, conference]{ieeeconf}  

\IEEEoverridecommandlockouts                              

\usepackage{graphics} 
\usepackage{graphicx}

\usepackage{epsfig} 
\usepackage{amsmath} 
\usepackage{amssymb}  
\usepackage{listings}
\usepackage{colortbl}
\usepackage{adjustbox}
\usepackage{bm}
\usepackage{mathtools}

\usepackage{algorithm}
\usepackage[noend]{algpseudocode} 
\algnewcommand\AAND{\textbf{ and }}
\algnewcommand\Or{\textbf{ or }}

\usepackage{color}
\usepackage{citesort}
\usepackage{flushend}
\usepackage{url}
\usepackage[breaklinks]{hyperref}
\usepackage[capitalise]{cleveref}
\usepackage{booktabs}
\usepackage{makecell}
\usepackage{siunitx}
\usepackage[nolist,nohyperlinks]{acronym}
\usepackage{listings}
\usepackage[table, dvipsnames]{xcolor} 

\definecolor{darkgreen}{RGB}{0, 100, 0}
\definecolor{darkred}{RGB}{200, 0, 0}

\lstdefinestyle{pythonstyle}{
    language=Python,
    basicstyle=\ttfamily\small,  
    keywordstyle=\bfseries\color{darkred},  
    keywordstyle={[2]\color{red}},  
    commentstyle=\color{darkgreen}\itshape,  
    stringstyle=\color{red},  
    showstringspaces=false,  
    numbers=none,  
    breaklines=true,  
    frame=none,  
    tabsize=4,  
    morekeywords={self},  
    morekeywords={[2]def},  
}
\usepackage{multirow} 

\acrodef{method}[AOM]{ACRONYM OF METHOD}
\acrodef{gnss}[GNSS]{Global Navigation Satellite System}
\acrodef{ransac}[RANSAC]{Random Sample Consensus}
\acrodef{slam}[SLAM]{Simultaneous Localization And Mapping}
\acrodef{pca}[PCA]{Principal Component Analysis}
\acrodef{ekf}[EKF]{Extended Kalman Filter}
\acrodef{rmse}[RMSE]{Root Mean Square Error} 
\acrodef{ape}[APE]{Absolute Pose Error}
\acrodef{cfar}[CFAR]{Constant False Alarm Rate}
\acrodef{snr}[SNR]{Signal to Noise Ratio}
\acrodef{rcs}[RCS]{Radar Cross Section}
\acrodef{imu}[IMU]{Inertial Measurement Unit}
\acrodef{sgm}[SGM]{Segmi-Global Matching}
\acrodef{dnn}[DNN]{Deep Neural Network}
\acrodef{gru}[GRU]{Gated Recurrent Unit}
\acrodef{hpr}[HPR]{Hidden Point Removal}
\acrodef{raft}[RAFT]{Recurrent All-Pairs Field Transforms}
\acrodef{fov}[FOV]{Field of View}
\acrodef{mclab}[MC-lab]{Marine Cybernetics laboratory}
\acrodef{vio}[VIO]{Visual-Inertial Odometry}
\acrodef{rcm}[RCM]{Refractive Camera Model}
\acrodef{sfm}[SFM]{Structure from Motion}
\acrodef{sota}[SOTA]{State-of-the-Art}
\acrodef{mhsa}[MHSA]{Multi-Head Self-Attention}
\acrodef{dat}[DAT]{Deformable Attention Transformer}
\acrodef{fpn}[FPN]{Feature Pyramid Network}

\usepackage{tikz}
\usetikzlibrary{positioning, shapes.geometric, arrows.meta, decorations.pathreplacing, calligraphy}

\DeclareMathAlphabet{\pazocal}{OMS}{zplm}{m}{n}

\DeclareMathAlphabet{\mathpzc}{OT1}{pzc}{m}{it}

\usepackage{array}
\newcolumntype{C}[1]{>{\centering\arraybackslash}p{#1}}
\newcolumntype{M}[1]{>{\raggedright\arraybackslash}p{#1}}
\newcolumntype{L}[1]{>{\raggedright\let\newline\\\arraybackslash\hspace{0pt}}m{#1}}	
\newcolumntype{S}[1]{>{\centering\let\newline\\\arraybackslash\hspace{0pt}}m{#1}}
\newcolumntype{R}[1]{>{\raggedleft\let\newline\\\arraybackslash\hspace{0pt}}m{#1}}

\definecolor{light-yellow}{RGB}{245, 245, 220}
\definecolor{light-blue}{RGB}{173, 216, 230}

\makeatletter
\renewcommand*{\@opargbegintheorem}[3]{\trivlist
  \item[\hskip \labelsep{\itshape #1\ #2}] \textit{(#3)}\ }
\makeatother

\title{\LARGE \bf
Diffusion-based RGB-D Semantic Segmentation with Deformable Attention Transformer
}

\author{Minh Bui and Kostas Alexis
\thanks{This material was supported by the Research Council of Norway - Award NO-338694, and b) the European Commission - Grant No. 101121321.}
\thanks{The authors are with the Norwegian University of Science and Technology (NTNU), O. S. Bragstads Plass 2D, 7034, Trondheim, Norway {\tt\small minh.q.bui@ntnu.no}}
}

\begin{document}

\maketitle
\thispagestyle{empty}
\pagestyle{empty}

\begin{abstract}
Vision-based perception is of great importance for scene understanding in autonomous systems. RGB-D images are commonly used to capture both semantic and geometric features, but reliable interpretation is challenging due to unavoidable noise in real-world data. In this work, we introduce a diffusion-based framework to address the RGB-D semantic segmentation problem. Additionally, we demonstrate that utilizing a Deformable Attention Transformer as the encoder to extract features from depth images effectively captures the characteristics of invalid regions in depth measurements. Our generative framework shows a greater capacity to model the underlying distribution of RGB-D images, achieving robust performance in challenging scenarios with significantly less training time compared to discriminative methods. Experimental results indicate that our approach achieves State-of-the-Art performance on both the NYUv2 and SUN-RGBD datasets in general and especially in the most challenging of their image data. To demonstrate the practicality of our method, a real-world experiment is conducted to inspect an office and generate its 3D semantic map. Our project page will be available at \href{https://diffusionmms.github.io/}{https://diffusionmms.github.io/}
\end{abstract}

\section{Introduction}

 Semantic segmentation represents the challenging problem of assigning a class label to each pixel of an image and corresponds to an essential step in visual scene understanding. Linking the correct subsets of pixels to the correct class, ensuring that pixels around a certain object are not erroneously linked to its class, and dealing with challenging textures or light conditions represent persistent challenges in computer vision and robotics. Accordingly, a body of work has focused on this problem, with the most prominent and high-performance methods currently relying on deep learning techniques with diverse architectures~\cite{jia2024geminifusion,jiang2018rednet,zhao2017pyramid,ronneberger2015u,DFormer,hu2019acnet,chen2021spatial,wu2023hidanet}. Among the most recent approaches to the problem is that of combining multimodal sensor data with the hope of achieving improved overall performance and robustness. 

Combining information from various sensors equips robots with more resilient capabilities while working in complex environments since each modality can complement the weaknesses of the others. Current multimodal approaches for semantic segmentation typically use a simple dual-branch encoder-decoder architecture, with one branch being used for feature extraction from the RGB modality and the other for auxiliary modality feature extraction. 
\begin{figure}[h]
  \includegraphics[width=0.98\columnwidth]{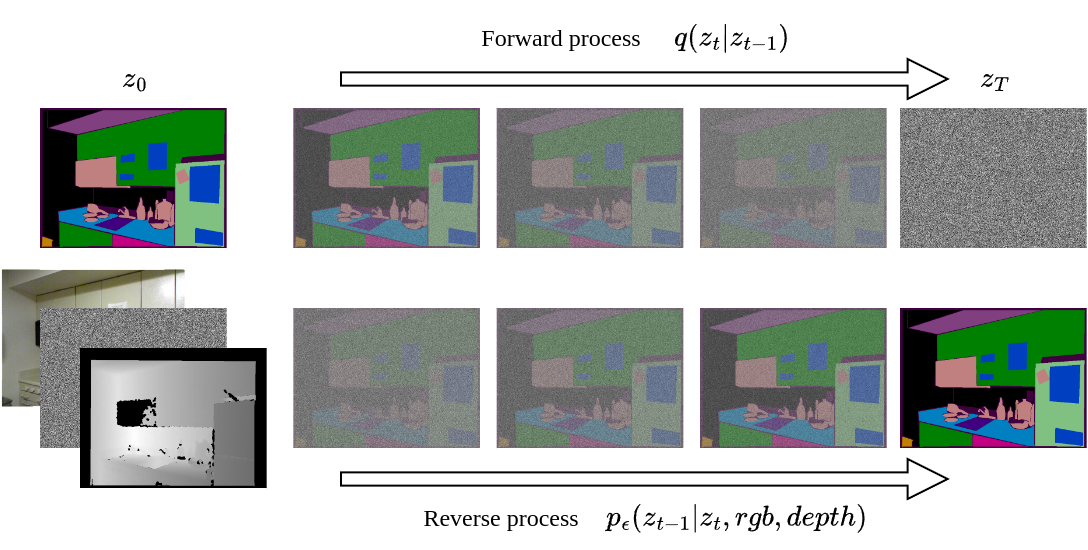}
  \caption{We formulate the RGB-D semantic segmentation task as a denoising diffusion process conditioned by RGB and depth images.}
      \label{fig:intro}
\vspace{-5mm}
\end{figure}
The segmentation mask is then achieved by merging the multimodal information using sophisticated fusion approaches. There are two main fusion schemes that are popular in recent works: interaction-based fusion \cite{zhang2023cmx,DFormer} and exchange-based fusion \cite{jia2024geminifusion,TokenFusion}.

Despite being straightforward, these kinds of methods often suffer performance degradation due to many invalid measurements from depth sensors (e.g., caused by stereo disparity errors, the effect of reflective surfaces, and other phenomena). To mitigate this, many methods exploit information from the corresponding RGB image to interpolate these missing values using the colorization method \cite{Levin2004}, or the HHA image algorithm \cite{gupta2014learning}. However, this step not only adds an additional computational cost but also compromises the integrity of how reality is captured. 

Furthermore, to the best of the authors' knowledge, all methods on RGB-D semantic segmentation only employ the discriminative paradigm \cite{DFormer,zhang2023cmx,jia2024geminifusion,TokenFusion}. Recently, generative-based methods such as diffusion models have been shown to achieve impressive results on several vision tasks. Although diffusion was originally designed for the image generation problem, many efforts have been made to apply it to RGB semantic segmentation \cite{baranchuk2022labelefficientsemanticsegmentationdiffusion,DDP,Generalist}. 

Motivated by its potential, we propose a simple yet effective diffusion framework for high-performance RGB-D semantic segmentation that addresses key domain challenges. In summary, our contributions are as follows: First, we demonstrate that using a deformable transformer as an image encoder in a discriminative-based architecture can alleviate the problem caused by invalid pixels in depth images. Second, we demonstrate that the use of diffusion can achieve improved results - compared to the \ac{sota} - combined with reduced training time. Third, experimental results demonstrate that our method achieves \ac{sota} on both the NYUv2 and SUNRGBD datasets and several challenging setups. We also validate our approach with a real-world drone experiment reconstructing a 3D semantic map of an office.

The remainder of this paper is organized as follows. Section~\ref{sec:related} presents related work, while the proposed method is detailed in Section~\ref{sec:method}. Evaluation studies are shown in Section~\ref{sec:evaluation}, while conclusions are drawn in Section~\ref{sec:concl}.

\section{Related work}\label{sec:related}

This work relates to the body of literature on RGB-D semantic segmentation alongside the works considering the role of diffusion in image segmentation in general. 

\subsection{RGB-D Semantic Segmentation}

An emerging trend for improving performance in RGB-D semantic segmentation is to create methods that enhance the 2D representation of RGB-D images. Designing complex fusion mechanisms to better utilize features extracted from both domains has become a de facto approach in the field. Most methods can be categorized into two main strategies. 

First, this relates to the interaction-based fusion strategy which focuses on integrating features from different modalities through direct interaction, typically via cross attention, or feature concatenation operations. The information from the two modalities can be merged directly at the input level through channel-wise averaging or concatenation and then processed by a single stream network as described in \cite{EarlyFusion1,EarlyFusion2,averaging,concatenate}. Despite its simplicity, merging raw data from different modalities too early can lead to a loss of modality-specific features, limited cross-modal interactions, increased sensitivity to noise and incomplete data. This approach lacks adaptability, contextual understanding, and often results in suboptimal performance, especially in complex scenarios. Another set of works proposed a fusion mechanism at the feature level which combines information from different modalities after extracting rich features separately from each modality. Zhang et al \cite{zhang2023cmx} introduced a feature rectification module to refine features between modalities and a feature fusion module to enable the comprehensive exchange of long-range contexts before final fusion. Ying et al \cite{UCTNet} suggest that using depth map uncertainty as an auxiliary signal can enhance segmentation accuracy and robustness. Yin et al \cite{DFormer} suggest an attention-based fusion scheme and performing supervised RGB-D pre-training on ImageNet-1K to develop a more effective backbone network for downstream tasks.

Second, the exchange-based fusion strategy involves the exchange of information between modalities through shared representations. The idea is to dynamically refine features from each modality based on insights gained from the other. TokenFusion \cite{TokenFusion} employs the prune-then-substitute scheme to replace uninformative tokens with more valuable ones from the other modality. This work is studied further by Jia et al \cite{jia2024geminifusion}, where it is suggested that the risk of all tokens engaging in unnecessary exchange can lead to significant information loss, and proposing an exchange-based strategy based on cross-modal transformers instead. 


\subsection{Diffusion in Image Segmentation}

All current literature in the field of RGB-D semantic segmentation follows the discriminative paradigm which broadly represents the community standard for semantic segmentation. However, generative models have recently taken the community by storm with their remarkable performance in the image generation task. Several studies suggest that generative models can achieve promising results on segmentation tasks \cite{DDP,Generalist,baranchuk2022labelefficientsemanticsegmentationdiffusion}. Baranchuk et al \cite{baranchuk2022labelefficientsemanticsegmentationdiffusion} illustrated that feature representations learned from pre-trained diffusion models can be fine-tuned for the segmentation task and achieve \ac{sota} results with limited labels. In \cite{Generalist}, Chen et al demonstrated a general framework based on diffusion models for panoptic segmentation on images and videos. In \cite{DDP}, Ji et al proposed a diffusion-based architecture that is suitable for several visual dense prediction tasks, including semantic segmentation, depth estimation, and BEV map segmentation. These results are based on RGB image data. Their promising results in combination with the benefits of multi-modality motivated the incorporation of diffusion in the proposed architecture for RGB-D semantic segmentation.

\section{Proposed Method}\label{sec:method}
\subsection{Preliminaries}
\begin{figure*}[h]
\centering          
  \includegraphics[width=2.0\columnwidth]{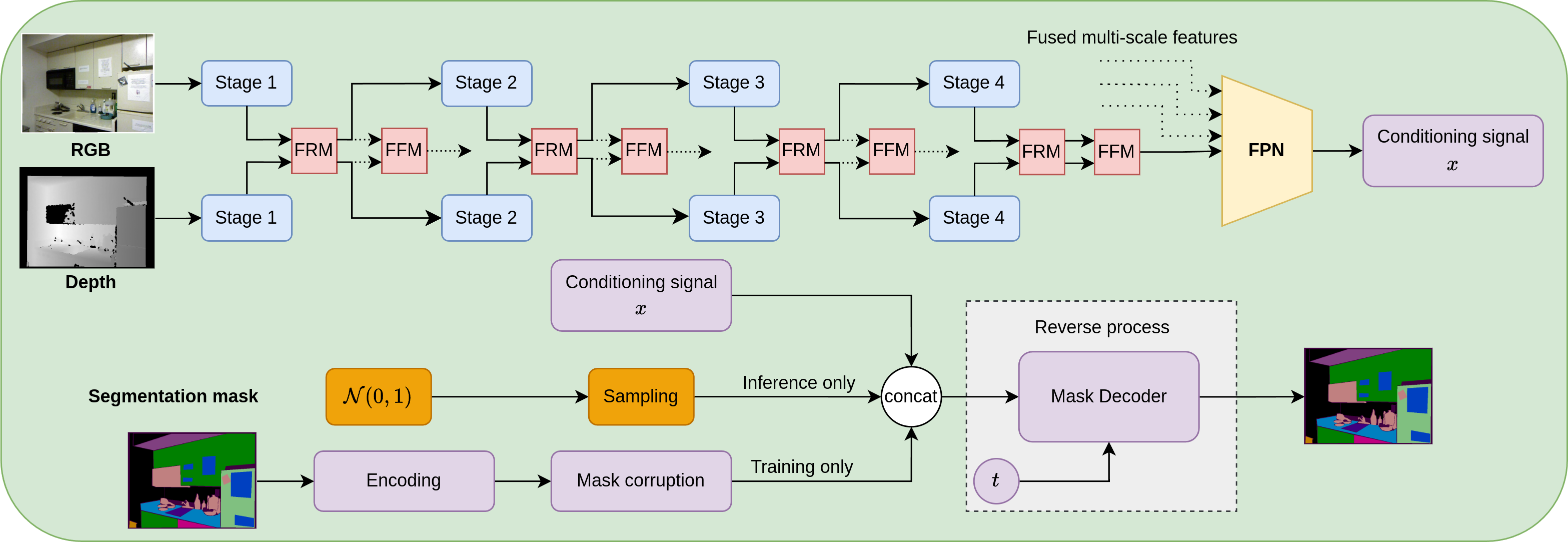}
  \caption{The architecture of our RGBD semantic mask generation framework. A Deformable Attention Transformer is used as the hierarchical encoder to extract features from RGB and depth images. Multi-scale features from both branches are then processed using fusion modules followed by a Feature Pyramid Network to create a conditioning signal $x$ that matches the shape of the noisy segmentation ground truth. A deformable attention mask encoder is trained to gradually denoise the concatenated signal to generate the segmentation mask.}
\label{fig:architecture}
\end{figure*}

\subsubsection{Deformable Attention Transformer}
We first revisit the vanilla Vision Transformer with the attention mechanism at its heart. Given an input feature vector $x \in \mathbb{R}^{N\times C}$, a M heads \ac{mhsa} block is defined as
\begin{eqnarray}
    q &=& W_qx,~ k=W_kx,~ v = W_vx \\
    z_m &=& \textrm{softmax}(q_m k_m^{\top}/\sqrt{d}) v_m \\ 
    z &=& \textrm{concat}(z_1,..., z_M)W_{out}
\end{eqnarray} 
where $W_{out}, W_q, W_k, W_v$ are projection matrices, $d$ represents the dimension of each head, $z_m$ represents the output embedding of the $m$-th attention head, and $q_m, k_m, v_m \in \mathbb{R}^{N \times d}$ represent query, key, and value embeddings respectively.

Many attempts have been made to address the quadratic complexity with respect to input dimension in the vanilla Vision Transformer \cite{SwinTransformer,PVT,DAT,DAT++}. The Deformable Attention Transformer \cite{DAT,DAT++} introduces the idea of learning a few sets of sampling offsets, shared across all queries, to adjust keys and values to important regions, based on the observation that global attention typically produces similar patterns for various queries \cite{cao2019gcnetnonlocalnetworksmeet,zhou2021deepvitdeepervisiontransformer}. By doing that, it effectively captures relationships between tokens by focusing on key areas of the feature map. Given the initial attention region position $p$, it is dynamically updated through deformable sampling points learned from queries via offset networks $\Delta p =\epsilon_{offset}(q)$. The features are sampled at the locations of deformed points via a bilinear interpolation function $\varphi$. They serve as keys and values, transformed by projection matrices.
\begin{eqnarray}
        q &=& W_qx, \Bar{k}=W_k\Bar{x}, \Bar{v} = W_v\Bar{x}\\
        \Delta p &=&\epsilon_{\textrm{offset}}(q), \Bar{x} = \varphi(x; p + \Delta p)
\end{eqnarray} 
where $\Bar{k}, \Bar{v}$ respectively denotes the deformed key and value embeddings. More details can be found in \cite{DAT,DAT++}.

\subsubsection{Diffusion models}
Diffusion models \cite{DDPM,DDIM} are generative models that can learn the underlying data distribution, allowing one to synthesize new data points from pure noise. They consist of two processes. In the forward process, noise is iteratively added to the data sample $z_0$, converting it into a latent noisy sample $z_t$ based on a noise scheduler $\beta_s$ \cite{DDIM,DDPM}. The whole process is mathematically defined as 

\begin{equation}
    q(z_t | z_0) = \mathcal{N}(z_t; \sqrt{\Bar{\alpha}_t}z_0, (1-\Bar{\alpha}_tI)), t \in {0,1,..., T}
    \label{eqn: forward diffusion}
\end{equation}
where $\Bar{\alpha}_t = \prod_{s=0}^{t}(1 - \beta_s)$, $I$ is the identity matrix.

One can control the output of diffusion models to generate samples belonging to a domain of interest simply by concatenating a conditioning signal $x$ to the noisy sample $z_t$.
In the training stage, a neural network $f_{\epsilon}(z_t, x, t)$ is trained to predict $z_0$ from $z_t$ given the conditioning signal $x$ by optimizing a training objective function. In the inference stage, starting from a sample of noise $z_T$, the data sample $z_0$ is generated by applying the model $f_{\epsilon}$ iteratively with transition rules such as ones described in \cite{DDPM,DDIM}. 

\subsection{Proposed architecture}

\begin{figure}[h]
    \centering
    \includegraphics[width=0.98\columnwidth]{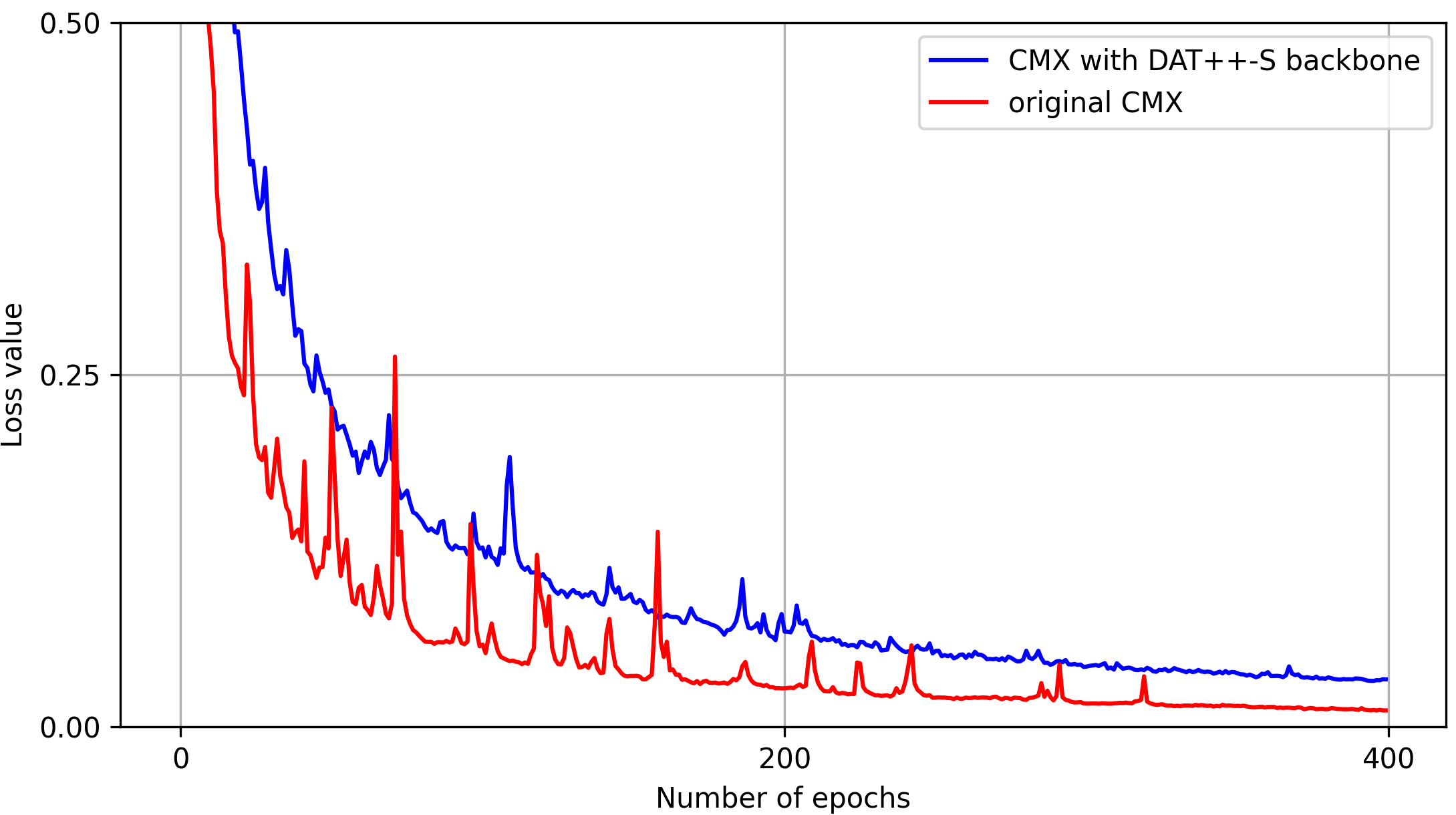}
    \caption{Loss comparison between training the original CMX~\cite{zhang2023cmx} model and when using DAT++-S as the encoder for raw depth images.}
    \label{fig:loss}
\end{figure}

We start by considering the method described in \cite{zhang2023cmx} as the baseline. In \cite{zhang2023cmx}, Zhang et al proposed a sophisticated fusion module to facilitate interactions between RGB and depth images. However, in order to achieve good results, it relies on the three-channel HHA encoding \cite{gupta2014learning} of the depth images. To understand the underlying challenges of the method, we retrained their models on raw depth images with a high number of invalid pixels which can be up to over 80\% the total number of pixels in a depth image and found that the training loss exhibits multiple spikes (see Figure \ref{fig:loss}) suggesting training instability and possible overfitting to the noise in the depth images. We argue that instead utilizing the \ac{dat} as the encoder -- with its characteristic of having the adaptive spatial aggregation conditioned by input and task information \cite{InternImage}-- is well suited to tackle the challenge posed by the invalid pixels in depth images. Furthermore, a diffusion model, as a denoising process, is advantageous when learning the underlying distribution of depth images given their nature of having uncertain measurements. 

Accordingly, we formulate the RGB-D semantic segmentation task as a conditioning image generation task. The goal is to learn the underlying distribution of the segmentation mask conditioned by RGB and depth images. In inference time, noise sampled from a normal distribution is concatenated with fusion features extracted from RGB-D inputs via a deformable attention transformer used as an encoder. The combined signal then goes through the reverse process to iteratively generate the final segmentation mask given the RGB-D inputs. The overall architecture of our method is illustrated in Figure \ref{fig:architecture}. In the following sections, the details of each component in the architecture are presented.

\subsubsection{Encoder}
The double encoder architecture is used to obtain the conditioning signal from paired RGB-D frames. We employ the \ac{dat} described in \cite{DAT++} as the encoder to extract multi-scale features at different resolutions from both modalities. We use the feature rectification modules (FRM) and feature fusion module (FFM) described in \cite{zhang2023cmx} to obtain multi-scale RGB-D fusion features. Notice that the output of the conditioning signal should be of the same size as the segmentation ground-truth. To reduce computation cost for the mask decoder, we resize the segmentation mask from $h \times w$ to $\frac{h}{4} \times \frac{w}{4}$. Accordingly, we generate the conditioning signal of shape $256 \times \frac{h}{4} \times \frac{w}{4}$ by merging multi-scale RGB-D fusion features via a \ac{fpn} and then aggregating the output using a $1 \times 1$ convolution. 

\subsubsection{Mask Decoder}
The mask decoder $f_{\epsilon}$ takes the noisy mask $y_t$ and the conditioning signal via concatenation as input. It is then trained to reconstruct the segmentation ground truth with the original size using the standard cross-entropy loss for the semantic segmentation task. Following \cite{DDP}, we form the mask decoder by stacking six layers of deformable attention blocks \cite{DeformableDETR} with time embedding.

\subsubsection{Training}
During training, we first perform the forward process which converts the segmentation label $y$ into the noisy map $y_t$ and then train the reverse model to learn how to remove noise. The training procedure is outlined in Algorithm \ref{alg: training}. Details about components in the forward process are presented below:

\textbf{Label encoding.} Diffusion models were originally designed to work with continuous data and Gaussian noise. Several studies have investigated ways to apply diffusion to tasks with discrete labels \cite{AnalogBit,Generalist,DDP} such as semantic segmentation. Inspired by the work of Ji et al \cite{DDP}, we use the class embedding approach in which a learnable embedding layer is used to project discrete labels into a high-dimensional space, normalized by a sigmoid function. The encoded labels are scaled to the range of $[-s, s]$ as shown in Algorithm \ref{alg: training}.

\textbf{Mask Corruption.}  The Gaussian noise is added to the encoded segmentation ground truth to produce the noise mask $y_t$. The magnitude of noise to be added is regulated by $\alpha_t$, which follows a decreasing pattern over the timesteps $t \in [0, 1]$. Various noise schedules, such as cosine \cite{nichol2021improveddenoisingdiffusionprobabilistic} and linear schedules \cite{DDPM}, are analyzed in Section \ref{sec:evaluation}.

\begin{algorithm}
\caption{Training procedure}\label{euclid}
\begin{lstlisting}[style=pythonstyle]
def train(rgb, depth, mask):
# obtain fused features
rgb_enc = rgb_encoder(rgb) #DAT encoder
depth_enc = depth_encoder(depth)#DAT encoder
fused = fusion(rgb_enc, depth_enc)# CMX based
# encode segmentation mask
mask_enc = encoding(mask) 
mask_enc = (sigmoid(mask_enc) * 2 - 1) * s
# noisify gt
t, eps = uniform(0, 1), normal(mean=0, std=1)
mask_noise = sqrt(alpha_bar(t)) * mask_enc + sqrt(1 - alpha_bar(t))*eps
# predict and calculate loss
mask_pred = mask_decoder(mask_noise,fused,t)
loss = cross_entropy(mask_pred, mask)
return loss    
\end{lstlisting}
\label{alg: training}
\end{algorithm}
\subsubsection{Inference}
The inference process is outlined in Algorithm \ref{alg: sampling}. Given paired RGB and depth images as the conditional inputs, the model begins with a random noise map generated from a Gaussian distribution and progressively improves the prediction. To minimize the number of iterative steps, we choose the DDIM update rule \cite{DDIM,DDP} for the sampling process. The key hyperparameter in this update rule is the time difference \texttt{td} which determines how far apart consecutive timesteps are chosen during the reverse diffusion process. Larger time gaps between steps allow for faster sampling at the potential cost of sample quality. Our experiments show that \texttt{td}$~=1$ works best for our method. The details of the DDIM update rule are presented in Algorithm \ref{alg: DDIM update}.

\begin{algorithm}
\caption{Sampling procedure}\label{euclid}
\begin{lstlisting}[style=pythonstyle]  
def sampling(rgb, depth, steps, td=1):
# obtain fused features
rgb_enc = rgb_encoder(rgb) #DAT encoder
depth_enc = depth_encoder(depth)#DAT encoder 
fused = fusion(rgb_enc, depth_enc)# CMX based
# sample noise
mask_t = normal(0, 1) 
for step in range(steps):
 # get time intervals
 t_now = 1 - step / steps
 t_next = max(1 - (step + 1 + td) / steps, 0)
 # predict mask_0 from mask_t
 mask_pred = mask_decoder(mask_t,fused,t_now)
 # estimate mask_t at t_next
 mask_t = ddim_step(mask_t, mask_pred, t_now, t_next)
return mask_pred
    \end{lstlisting}
    \label{alg: sampling}
\end{algorithm}
\begin{algorithm}
\caption{DDIM update rule}\label{euclid}
\begin{lstlisting}[style=pythonstyle][escapeinside={(*}{*)}]
def alpha_bar(t, ns=0.0002, ds=0.00025):
"""cosine noise scheduler"""
n = torch.cos((t+ns)/(1+ds) * math.pi/2)**-2
return -torch.log(n - 1, eps=1e-5)

def ddim_step(mask_t,mask_pred,t_now,t_next):
"""estimate x with DDIM update rule."""
(*@$\alpha_{now}$ @*) = alpha_bar(t_now)
(*@$\alpha_{next}$ @*) = alpha_bar(t_next)
mask_enc = encoding(mask_pred)
mask_enc = (sigmoid(mask_enc) * 2 - 1) * s
eps = (*@$\frac{1}{\sqrt{1- \alpha_{now}}}$ @*)*(mask_t -(*@ $\sqrt{\alpha_{now}}$ @*)* mask_enc)
mask_next =(*@ $\sqrt{\alpha_{next}}$ @*)* mask_enc +(*@ $\sqrt{1 - \alpha_{now}}$ @*)*eps
return mask_next
\end{lstlisting}
    \label{alg: DDIM update}
\end{algorithm}

\section{Experimental Evaluation}\label{sec:evaluation}
\subsection{Training setup}

\subsubsection{Datasets and Metrics} We test our method on the validation sets of two popular RGB-D datasets: NYUv2 \cite{NYUv2} and SUN-RGBD \cite{SUNRGBD}. For the NYUv2 dataset, which has 40 labeled classes, we follow the standard split of 795 training and 654 testing images. The SUN-RGBD dataset is seven times larger than the NYUv2 dataset in size, containing 5285 training and 5050 testing images with 37 labeled classes. To evaluate the performance of our method, we use the standard mean intersection over union (meanIoU) metric \cite{meanIOU} for the semantic segmentation task.
\begin{figure*}[htbp]
    \centering
    \includegraphics[width=2.0\columnwidth]{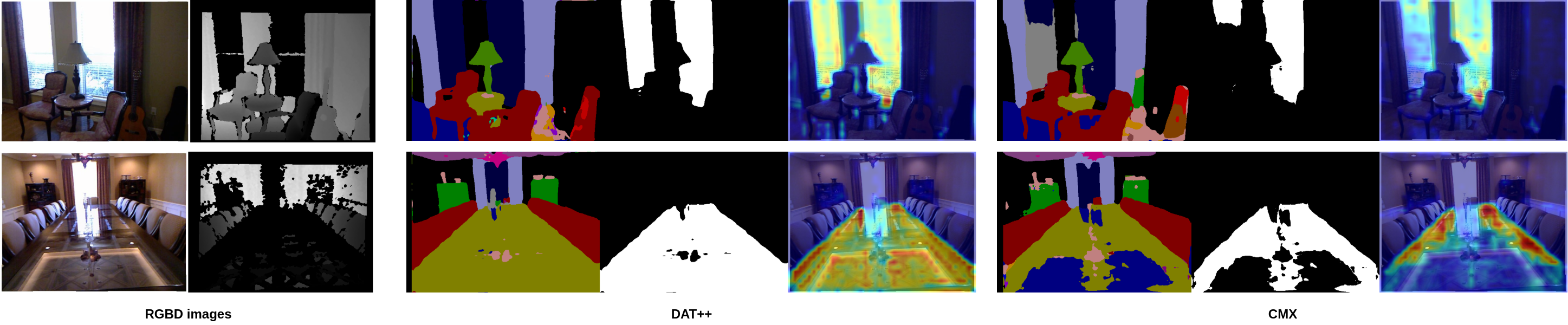}
    \caption{GradCAM (Gradient-weighted Class Activation Map) heatmaps visualizing class activations from the final layer of the backbone used in our method and in CMX \cite{zhang2023cmx}. For each method, we present (from left to right) the predicted semantic segmentation mask, the binary mask for the selected class, and the corresponding GradCAM heatmap. The top row displays results for the class 'window'. The bottom row shows results for the class 'table'.}
    \vspace{-2ex}
    \label{fig: gradcam}
\end{figure*}
\subsubsection{Training details and hyperparameters}
For all experiments, we train our models using the AdamW optimizer \cite{AdamW}, with an initial learning rate of $6 \times 10^{-5}$ and a weight decay of 0.01. The learning rate is regulated using the warmup polynomial decay scheduler. As seen in many previous works ~\cite{zhang2023cmx,DFormer,UCTNet}, the discriminative paradigm needs at least 300 to 500 epochs to obtain good results on both NYUv2 and SUN-RGBD datasets. We found that our diffusion-based model can converge in a significantly less amount of time. We train the NYUv2 and SUN-RGBD datasets for 100 and 50 epochs, respectively. For data augmentation, we perform resize, random crop, and random flip operations on both RGB and depth images.
\subsection{Ablation study}
\subsubsection{Comparison with the baseline CMX~\cite{zhang2023cmx}}
We adopt the fusion mechanism from CMX~\cite{zhang2023cmx} as the foundation of our architecture, but introduce two key enhancements to improve performance on depth images with invalid regions. First, we replace encoders used in CMX~\cite{zhang2023cmx} with a Deformable Attention Transformer encoder that adaptively attends to spatially relevant regions in both RGB and depth images. Second, we shift the training paradigm for semantic segmentation from a discriminative approach to a generative one. 

To visualize the effectiveness of the Deformable Attention Transformer encoder in obtaining more robust features of depth images, we use the Gradient-weighted Class Activation Mapping (GradCAM) \cite{GradCAM} technique to highlight the regions in an input image that are most influential for a model’s decision. As shown in Figure 4, the Deformable Attention Transformer encoder can handle invalid regions caused by reflective surfaces such as windows or glass better than encoders used in CMX~\cite{zhang2023cmx}. The superiority of \ac{dat} on RGB-D semantic segmentation over other non data-dependent encoders such as MixTransformer \cite{SegFormer} is quantitatively shown in Table \ref{tab: DAT vs MiT}. We take the result reported in \cite{zhang2023cmx} as the baseline. By simply changing the encoder of the model to DAT++~\cite{DAT++}, we achieved a significant boost in terms of the meanIoU metric in the NYUv2 dataset, as seen in Table \ref{tab: DAT vs MiT} across all versions of the DAT++ backbone. Note that the number of parameters reported in Table \ref{tab: DAT vs MiT} refers to the size of the backbone network used for feature extraction from each input. Utilizing the tiny version of DAT++ achieves a  $1.8\%$ improvement in mean IoU while reducing the number of parameters to one-third. Note that the number of parameters in Table \ref{tab: DAT vs MiT} reflects the size of the encoder used to extract features from each modality. However, when evaluating a larger dataset like SUN-RGBD using the discriminative model, we did not achieve any performance boost. By using the diffusion model as the alternating paradigm, we achieved better results compared to the discriminative model on both datasets, as seen in Table \ref{tab: DAT vs MiT} while spending significantly less time/epochs to train.

\begin{figure}[htbp]
    \centering
    \includegraphics[width=0.95\columnwidth]{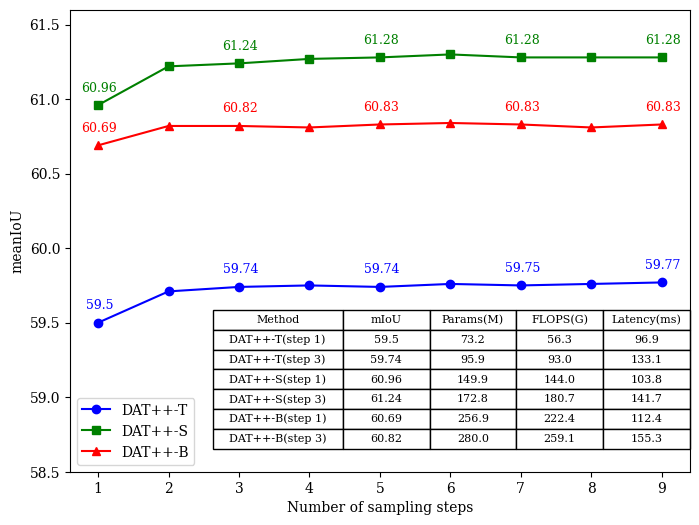}
    \caption{Performance and computational cost of our method with different numbers of sampling steps used for inference on NYUv2 dataset}
    \label{fig:number of step ablation study}
\end{figure}

Our method inherits the multiple-step inference procedure from the diffusion process. This introduces a tradeoff between model performance and computational cost, depending on the number of sampling steps required for inference. Typically, at least 50 sampling steps are necessary to achieve high-fidelity results in image generation tasks \cite{DDIM}. However, as shown in Figure \ref{fig:number of step ablation study}, the performance of the diffusion process for the semantic segmentation task quickly saturates after just a few sampling steps. Even with a single sampling step, our method outperforms the baseline by a significant margin. The reported latency of our models was measured using an NVIDIA GeForce RTX 3090.

\begin{table}[h]
  \centering
  \caption{Performance comparison between DAT++ and MiT-B5 using different paradigms on NYUv2 and SUN-RGBD datasets. The number of parameters refers to the size of the encoder used to extract features from each modality.}
  \label{tab: DAT vs MiT}
      \begin{adjustbox}{width=0.48\textwidth}

\begin{tabular}{|c|c|c|c|}
    \hline
    Methods & Params(M) & NYUv2 & SUN-RGBD\\
    \hline
    CMX (MiT-B5)  & 82 & 56.9 & 52.4 \\
    Discriminative-based (DAT++-T) & 24 & 58.7 & 51.7\\ 
    Discriminative-based (DAT++-S) & 53 & 60.2 & 52.4\\ 
    Discriminative-based (DAT++-B)& 93 & 59.9  & 51.8\\ 
     Diffusion-based (DAT++-T) & 24 & 59.7  & 52.6\\
     Diffusion-based (DAT++-S) & 53 & \textbf{61.5}  & 53.8\\
     Diffusion-based (DAT++-B) & 93 & 60.8  & \textbf{54.0}\\
\hline
\end{tabular}
  \end{adjustbox}

\end{table}

\subsubsection{Evaluation results on challenging datasets}
We evaluate our diffusion based framework using 3 encoder variants, namely DAT++-T, DAT++-S, DAT++-B \cite{DAT++}. They achieve \ac{sota} results compared to current literature in the field, as shown in Table \ref{tab: SOTA comparison}. On the NYUv2 dataset, our method with the  DAT++-S encoder achieves the best meanIoU of 61.2\% which is the number one on the dataset leaderboard at the time of writing. Interestingly, increasing the model size to DAT++-B leads to 0.4\% performance drop. This might be the overfitting problem when using a large model on a small dataset like the NYUv2 dataset. On the SUN-RGBD dataset, we achieved the best result of 53.9\% with the DAT++-B encoder. We lose only to GeminiFusion with the Swin-Large encoder (369.2M parameters), which is much larger than our biggest model with the DAT++-B encoder (256.8M parameters). Therefore, our method remains the best method performing on the SUN-RGBD dataset compared with other similar size. Our method outperforms others more on the small NYUv2 dataset than on the larger SUN-RGBD, showing its strength with limited data. Furthermore, when we evaluate on the SUN-RGBD dataset using the model trained on the NYUv2 dataset and vice versa, we observe only a modest performance drop for all variants of our model, which suggests a good generalization capability of our method.

\begin{table}[h]
  \centering
\arrayrulecolor{black} 
  \caption{Comparison with other State-of-the-art methods on the NYUv2 and SUN-RGBD datasets. Red numbers indicate the results when evaluating the SUN-RGBD dataset using the model trained on NYUv2 and vice versa. The scale value $s = 0.01$}
  \label{tab: SOTA comparison}
    \begin{adjustbox}{width=0.5\textwidth}
\begin{tabular}{|c|c|c|c|c|}
    \hline
     Methods & Backbone & Params(M) & NYUv2 & SUN-RGBD  \\
    \hline
    CMX (MiT-B4)\cite{zhang2023cmx} & MiT-B4 & 139.9 & 56.3 & 52.1 \\
    CMX (MiT-B5)\cite{zhang2023cmx} & MiT-B5 & 181.1 & 56.9 & 52.4 \\
    DFormer-S\cite{DFormer} & DFormer-S & 18.7 & 53.6 & 50.0 \\
    DFormer-B\cite{DFormer}& DFormer-B &  29.5& 56.9 & 51.2 \\
    DFormer-L\cite{DFormer}& DFormer-L &  39.0& 57.2 & 52.5 \\
    TokenFusion\cite{TokenFusion} &MiT-B3 & 45.9 & 54.2 & 52.4 \\
    GeminiFusion \cite{jia2024geminifusion} & MiT-B3 &75.8 & 56.8 & 52.7\\ 
    GeminiFusion \cite{jia2024geminifusion} & MiT-B5 & 137.2& 57.7 & 53.3\\ 
    GeminiFusion \cite{jia2024geminifusion} & Swin-L &369.2 & 60.9 & \textbf{\textit{54.6}}\\ 
    \cellcolor{light-yellow}\textbf{Ours (step 3)}  &\cellcolor{light-yellow}DAT++-T &\cellcolor{light-yellow} 73.2 & \cellcolor{light-yellow}95.9 (\textcolor{red}{59.7}) & \cellcolor{light-yellow}52.6 (\textcolor{red}{50.6})\\ 
    \cellcolor{light-yellow}\textbf{Ours (step 3)}  &\cellcolor{light-yellow}DAT++-S &\cellcolor{light-yellow} 172.8& \cellcolor{light-yellow}\textbf{61.2} (\textcolor{red}{61.1}) & \cellcolor{light-yellow}53.7 (\textcolor{red}{51.2})\\ 
    \cellcolor{light-yellow}\textbf{Ours (step 3)}  &\cellcolor{light-yellow}DAT++-B &\cellcolor{light-yellow} 280.0 & \cellcolor{light-yellow}60.8 (\textcolor{red}{60.4})& \cellcolor{light-yellow}\textbf{53.9} (\textcolor{red}{51.4})\\ 
\hline
\end{tabular}
  \end{adjustbox}
\end{table}

\begin{figure*}[h]
    \centering
    \includegraphics[width=2.0\columnwidth]{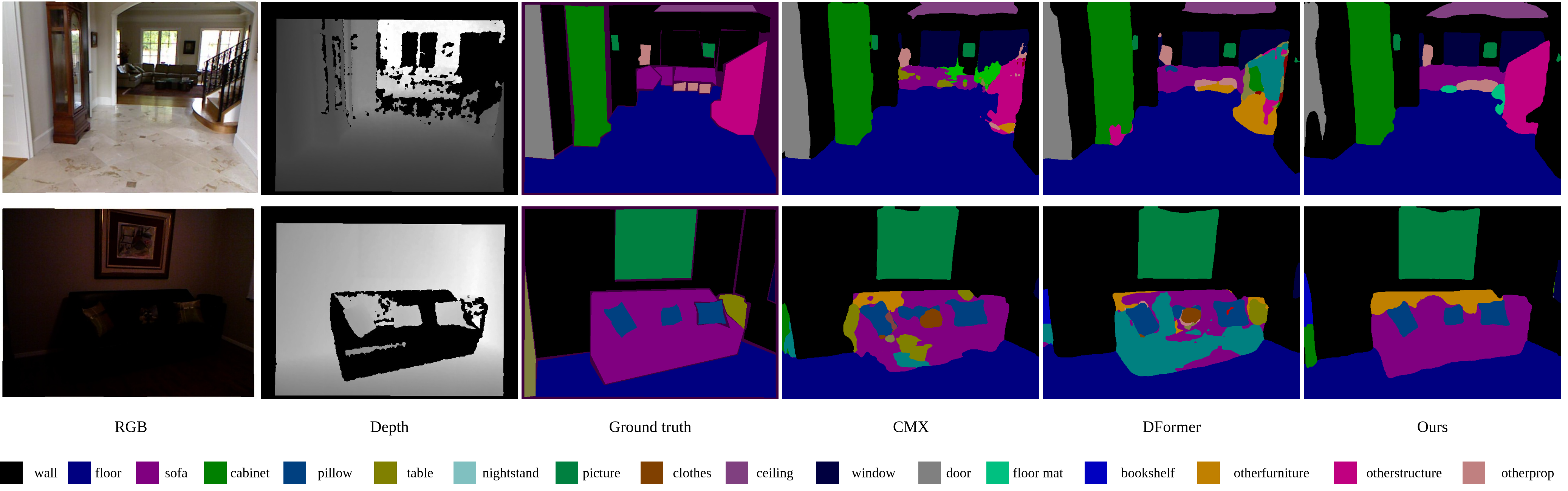}
    \caption{Our segmentation results show that our method has a better understanding of invalid depth regions than other methods in both normal (upper row) and low-light conditions (lower row)}
    \vspace{-2ex}
    \label{fig: segmentation results}
\end{figure*}

\begin{table*}[h]
  \centering
    \caption{Comparison with other methods on challenging subsets. The red numbers indicate the relative performance drop compared with evaluation results on the original datasets. DAT++-S is used as the encoder.}
  \label{tab: 3 datasets}
\arrayrulecolor{black} 
\begin{tabular}{|c|c|c|c|c|c|}
    \hline
    Datasets & Methods & Nominal & Low-light & Invalid & Small \\
    \hline
    \multirow{5}{*}{NYUv2}  &  CMX (MiT-B2)\cite{zhang2023cmx}  & 54.1 & 50.6 (\textcolor{red}{3.5}) & 52.8 (\textcolor{red}{1.3}) & 49.2 (\textcolor{red}{4.9}) \\
    &DELIVER\cite{zhang2023deliveringarbitrarymodalsemanticsegmentation}  & 56.3 & 53.8 (\textcolor{red}{2.5})& 51.8 (\textcolor{red}{4.5}) & 48.5 (\textcolor{red}{7.8})\\
    & DFormer-B\cite{DFormer} & 55.6 & 52.8 (\textcolor{red}{2.8})& 51.5 (\textcolor{red}{4.1})& 50.4 (\textcolor{red}{5.2})\\ 
    & DFormer-L\cite{DFormer} & 57.2 & 53.6 (\textcolor{red}{3.6})& 53.1 (\textcolor{red}{4.1})& 51.6 (\textcolor{red}{5.6})\\
    &\cellcolor{light-yellow}\textbf{Ours} (s=0.01) (step 3) &\cellcolor{light-yellow}61.2 &\cellcolor{light-yellow}58.8 (\textcolor{red}{2.4})&\cellcolor{light-yellow}59.3 (\textcolor{red}{1.9})&\cellcolor{light-yellow}56.6 (\textcolor{red}{4.6}) \\
    & \cellcolor{light-yellow}\textbf{Ours} (s=0.03) (step 3) &\cellcolor{light-yellow}\textbf{61.5} &\cellcolor{light-yellow}58.9 (\textcolor{red}{2.6})&\cellcolor{light-yellow}58.7 (\textcolor{red}{2.8})&\cellcolor{light-yellow}56.6 (\textcolor{red}{4.9}) \\
    \hline
    \multirow{3}{*}{SUN-RGBD} & DFormer-B & 51.2 & 46.4 (\textcolor{red}{4.8})& 43.2 (\textcolor{red}{8.0})& 43.8 (\textcolor{red}{8.0})\\ 
    & DFormer-L & 52.5 & 47.3 (\textcolor{red}{5.2})& 43.2 (\textcolor{red}{8.3}) & 45.1 (\textcolor{red}{7.4}) \\
    & \cellcolor{light-yellow}\textbf{Ours} (s=0.01) (step 3) & \cellcolor{light-yellow}{53.7} & \cellcolor{light-yellow}52.2 (\textcolor{red}{1.5})& \cellcolor{light-yellow}48.2 (\textcolor{red}{5.5})& \cellcolor{light-yellow}49.0 (\textcolor{red}{4.7}) \\
    & \cellcolor{light-yellow}\textbf{Ours} (s=0.03) (step 3)& \cellcolor{light-yellow}\textbf{53.8} & \cellcolor{light-yellow}52.3 (\textcolor{red}{1.5})& \cellcolor{light-yellow}47.8 (\textcolor{red}{6.0})& \cellcolor{light-yellow}49.0 (\textcolor{red}{4.8}) \\

\hline
\end{tabular}
\end{table*}
\begin{table}[htp!]
  \centering
  \caption{Ablation study on different noise schedules with DAT++-S and scale value $s=0.01$.}
  \label{tab: linear vs cosine}
\begin{tabular}{|c|c|c|}
    \hline
    Scheduler & NYUv2 & SUN-RGBD  \\
    \hline
    cosine  & \textbf{61.2} & \textbf{53.7} \\
    linear & 60.9 & 53.6 \\ 
\hline
\end{tabular}
\end{table}

To further demonstrate the capacity of our method to exploit depth features under heavy uncertainty, we conduct further evaluations on three focused sub-datasets, namely 
\begin{itemize}
    \item \textbf{Most invalid pixels:} We sort the NYUv2 and SUN-RGBD datasets based on the percentage of invalid pixels in the depth images. We take the subset of the top 20\% depth images with the highest percentage of invalid pixels. We call this the `invalid' dataset.
    \item \textbf{Low light: } For each paired RGB-D frame in the original datasets, we deliberately decrease the intensity of the RGB images using a gamma correction operation. The intensity of a normalized image $I$ is adjusted based on the formula $I_{dark} = I^{\gamma}, \gamma = 2$. We call this the `low-light' subset of the dataset.
    \item \textbf{Small objects:} We ignore some labels in the ground truth that are less affected by noisy depth data and only evaluate the meanIoU metric of the rest. For NYUv2, we ignore the [``wall'', ``floor'', ``ceiling'', ``otherstructure'', ``otherfurniture'', ``otherprop''] labels. For SUN-RGBD, we ignore the [``wall'', ``floor'', ``ceiling''] labels. We call this the `small' objects dataset.
\end{itemize}
Table \ref{tab: 3 datasets} shows that our method outperforms others across all three challenging datasets, with less performance drop. This proves its effectiveness in modeling RGB-D images, even with a high percentage of invalid depth pixels. Qualitative results are shown in Figure \ref{fig: segmentation results}. The improved performance in visual degradation cases demonstrates our model’s ability to extract meaningful features directly from raw depth images, avoiding the need for interpolation used by other methods.
\subsubsection{Hyperparameter tuning} 
We conduct ablation experiments to find the best value for several key hyperparameters in diffusion models. Table \ref{tab: linear vs cosine} shows that using the cosine schedule achieves slightly better results than the linear schedule. Table \ref{tab: scale value} shows that 0.03 and 0.001 are the best scale values for NYUv2 and SUN-RGBD datasets, respectively. It is noted that $s=0.01$ and $s=0.03$ work well for both datasets, whereas $s=0.001$ improves the performance on SUN-RGBD by a small margin but disproportionately worsens the results on the NYUv2 dataset.
\begin{table}[htp!]
  \centering
  \caption{Ablation study on different scale values with DAT++-S}
  \label{tab: scale value}

\begin{tabular}{|c|c|c|}
    \hline
    Scale & NYUv2 & SUN-RGBD  \\
    \hline
    0.001  & 58.1 & \textbf{54.0} \\
    0.01  & 61.2 & 53.7 \\
    0.03  & \textbf{61.5} & 53.8  \\
    0.05  & 60.8 & 53.3  \\
    0.1  & 60.7 & 53.2 \\
\hline
\end{tabular}
\end{table}
\subsection{Volumetric mapping}
To further demonstrate the utility of our proposed method in a real-world scenario, we conducted an experiment in which a drone is tasked with inspecting an indoor office environment while building its semantic map. Our platform is equipped with an Intel RealSense D455 camera, a VectorNav VN-100 IMU, and an NVIDIA Orin NX 16 GB for onboard inference and mapping.
\begin{figure*}[htbp]
    \centering
    \includegraphics[width=2.0\columnwidth]{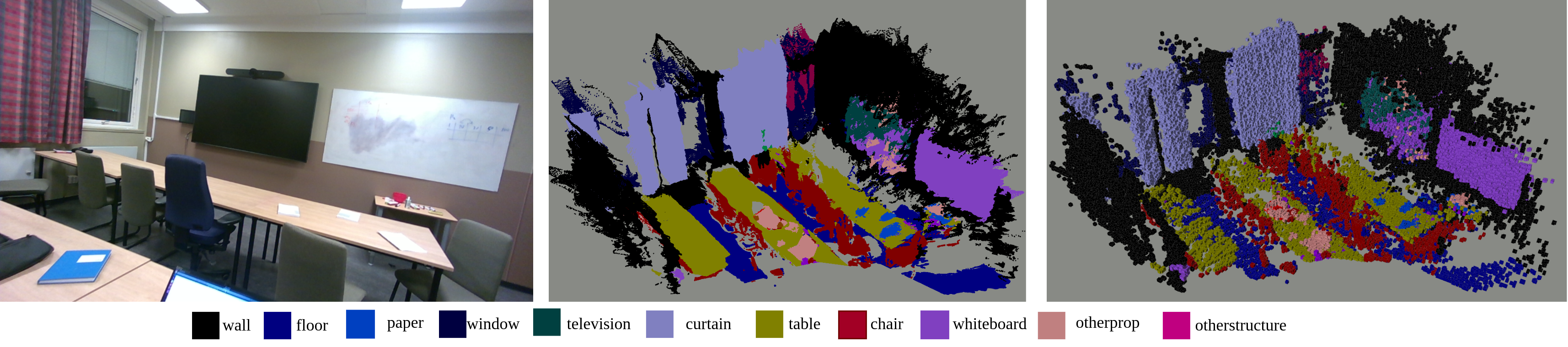}
    \caption{From left to right: RGB image of the office, the accumulated semantic point cloud generated by our model, and 3D semantic map reconstructed by Voxblox~\cite{voxblox} with voxel size of 2 cm. 'ceiling' voxel labels are removed for better visualization.}
    \vspace{-2ex}
    \label{fig: voxblox map}
\end{figure*}
The semantic segmentation is performed using our tiny version model with a single-step inference, running in 1.2 seconds per RGB-D frame on OrinNX. Dense 3D semantic reconstruction is performed by projecting 2D segmentation masks into 3D using depth measurements. The resulting semantic point cloud, along with camera poses estimated by ROVIO \cite{ROVIO}, is used to construct a voxel-based map with Voxblox \cite{voxblox}. During a 60s flight, we processed approximately 30 frames and fused them into a unified map. We follow \cite{Kimera} to build a vector of label probabilities for each bundle of rays during the bundled raycasting process and propagate semantic labels for each voxel traversed along the ray. For each voxel, its label probabilities are refined over time using a Bayesian update scheme \cite{SemanticFusion}, from which the most likely label is assigned. To reduce depth sensor errors near discontinuities and boundaries, we erode the 2D semantic mask to remove unreliable edge depths and filter isolated depth clusters for surface consistency. The resulting accumulated semantic point cloud and 3D map are shown in Figure \ref{fig: voxblox map}.

\section{Conclusion}\label{sec:concl}
Our diffusion-based framework enhances RGB-D semantic segmentation performance, using a Deformable Attention Transformer to robustly handle invalid depth regions. Experiments on NYUv2 and SUNRGBD achieve State-of-the-Art results with less training time, highlighting the potential of generative models for accurate vision reasoning in autonomous systems.





%


\bibliographystyle{IEEEtran}
\bibliography{references}

\end{document}